\documentclass[sigconf]{aamas}
\definecolor{skyblue}{HTML}{0395DE}
\definecolor{lightgray}{HTML}{999999}
\usepackage{subcaption}
\usepackage{amsmath}
\usepackage{multicol}
\usepackage[ruled,vlined, linesnumbered, boxed]{algorithm2e}
\newcommand{\model}{\textcolor{black}{Bayes-POMCP}}
\newcommand{\codeComment}[1]{\textcolor{lightgray}{#1}}

\newcommand{\old}[1]{\textcolor{black}{#1}}
\newcommand{\edit}[1]{\textcolor{black}{#1}}

\usepackage{balance} 
\usepackage{bbm}  

\setcopyright{ifaamas}
\acmConference[AAMAS '24]{Proc.\@ of the 23rd International Conference
on Autonomous Agents and Multiagent Systems (AAMAS 2024)}{May 6 -- 10, 2024}
{Auckland, New Zealand}{N.~Alechina, V.~Dignum, M.~Dastani, J.S.~Sichman (eds.)}
\copyrightyear{2024}
\acmYear{2024}
\acmDOI{}
\acmPrice{}
\acmISBN{}

\acmSubmissionID{995}

\title{Mixed-Initiative Human-Robot Teaming under Suboptimality with Online Bayesian Adaptation}

\author{Manisha Natarajan}
\authornote{\old{Both authors contributed equally to the paper.}}
\affiliation{
  \institution{Georgia Institute of Technology}
  \city{Atlanta, GA}
  \country{USA}}
\email{manisha.natarajan@cc.gatech.edu}

\author{Chunyue Xue}
\authornotemark[1]
\affiliation{
  \institution{Georgia Institute of Technology}
  \city{Atlanta, GA}
  \country{USA}}
\email{chunyuexue@gatech.edu}

\author{Sanne van Waveren}
\affiliation{
  \institution{Georgia Institute of Technology}
  \city{Atlanta, GA}
  \country{USA}}
\email{sanne@gatech.edu}

\author{Karen Feigh}
\affiliation{
  \institution{Georgia Institute of Technology}
  \city{Atlanta, GA}
  \country{USA}}
\email{karen.feigh@gatech.edu}

\author{Matthew Gombolay}
\affiliation{
  \institution{Georgia Institute of Technology}
  \city{Atlanta, GA}
  \country{USA}}
\email{matthew.gombolay@cc.gatech.edu}

\begin{abstract}
For effective human-agent teaming, robots and other artificial intelligence (AI) agents must infer their human partner’s abilities and behavioral response patterns and adapt accordingly. 
Most prior works make the unrealistic assumption that one or more teammates can act near-optimally. In real-world collaboration, humans and autonomous agents can be suboptimal, especially when each only has partial domain knowledge.
In this work, we develop computational modeling and optimization techniques for enhancing the performance of suboptimal human-agent teams, where the human and the agent \edit{have asymmetric capabilities} and act suboptimally due to incomplete environmental knowledge. We adopt an online Bayesian approach that enables a robot to infer people’s willingness to comply with its assistance in a sequential decision-making game. 
Our user studies show that user preferences and team performance indeed vary with robot intervention styles, and our approach for mixed-initiative collaborations enhances objective team performance ($p<.001$) and subjective measures, such as user's trust ($p<.001$) and perceived likeability of the robot ($p<.001$).
\end{abstract}


\newcommand{\BibTeX}{\rm B\kern-.05em{\sc i\kern-.025em b}\kern-.08em\TeX}

\begin{document}


\pagestyle{fancy}
\fancyhead{}


\maketitle 

\section{Introduction}

Human-agent teaming has the potential to leverage the unique capabilities of humans and artificial intelligence (AI) agents to enhance team performance. In real-world situations, both humans and agents can be suboptimal, especially when dealing with uncertainty \cite{human_suboptimality, kahn2017uncertainty}.
Imagine a human collaborating with a robot in an urban search-and-rescue (USAR) mission with reduced visibility due to fog or smoke. The human can take over control when the robot is more prone to make errors (e.g., in unstructured environments). 
Likewise, when human vision is limited, the robot can intervene or take control. 
To optimize this collaboration, robots need to develop a Theory of Mind \cite{rabinowitz2018machine}, i.e., the ability to infer the human teammates’ mental states and
anticipate their actions to determine when such intervention is beneficial.   
In this work, we look at \textit{mixed-initiative interactions}, where the robot actively models human behavior to decide when to intervene to maximize team performance.


In human-robot teams, mixed-initiative interaction refers to a collaborative strategy in which teammates opportunistically seize and relinquish initiative from and to each other during a mission, where initiative can range from low-level motion control to high-level goal specification~\cite{jiang2015mixed}. 
We study such interactions in a teaming task in which the human and the robot act suboptimally because they have partial knowledge of the environment. 
Specifically, the human teleoperates the robot, similar to USAR missions \cite{isaacs2022teleoperation}, and must collaborate with the robot (seize or relinquish control) to reach a goal location. 
The human and the robot have asymmetric capabilities \old{and non-identical, partial knowledge of the environment.}
\old{During the task, when the human selects an action, the robot can either comply with and execute the chosen action, interrupt by not executing the chosen action, or take control and execute an alternative action. If the robot interrupts or takes control, the human can decide whether to accept or oppose the robot's decision.}

Our goal is to learn a domain-agnostic robot policy that can effectively adapt to diverse users to maximize team performance without prior human interaction data.
Achieving such ad-hoc or zero-shot coordination with novel human partners has been a longstanding challenge in AI \cite{klien2004ten, paleja2021utility}. Recent works explore zero-shot human-AI collaboration by learning AI agent policies either from human-human demonstrations \cite{carroll2019utility, hong2023learning} or via self-play without any human data \cite{strouse2021collaborating, zhao2023maximum}. However, these approaches look at domains where humans and agents have symmetric capabilities. In contrast, our work delves into human-agent teaming with \emph{asymmetric capabilities}, \edit{where mixed-initiative teaming is essential. Prior works in mixed-initiative teaming have adopted strategies for switching control between humans and robots by estimating user performance \cite{chiou2021mixed} or operator engagement \cite{few2006improved}; our work differs by explicitly modeling user compliance to determine when robots should intervene}. 

Our contributions are two-fold. First, we propose a novel, online, Bayesian approach called \model, for zero-shot human-robot collaboration in mixed-initiative settings. We model the human-robot team as a Partially Observable Markov Decision Process (POMDP), \edit{where the robot maintains a belief over users' compliance tendencies}. 
Initially, the robot has high uncertainty about user preferences and willingness to comply. Through Bayesian Learning, the robot's estimation is iteratively refined, reducing its uncertainty upon subsequent interactions with the user. By conditioning the robot's policy on the uncertainty of the human model, 
our approach is more robust to adapt to a diverse pool of participants
than having a single, unified model for all subjects. To address the computational challenges in solving POMDPs and ensure that our approach is feasible to run online with novel users, \model{} employs a Monte-Carlo search (scalable to large state spaces) while anticipating appropriate user behavior with approximate belief updates. 

\old{Second, we design a new user study interface for examining mixed-initiative human-robot teaming. 
    We open-source our implementation\footnote{\edit{https://github.com/CORE-Robotics-Lab/Bayes-POMCP}}. We conduct two human-subjects experiments ($n=30$ and $n=28$) \old{with the interface} to show that (1) user preferences and team performance can vary when the robot employs different intervention styles, and (2) our proposed approach performs favorably on both objective (team performance) and subjective (users' trust, robot likeability) metrics with novel users.}


\section{Related Work}
\subsection{Modeling Human Behavior}
For seamless human-robot collaboration, robots must anticipate human behavior and act accordingly. Prior works have shown that robots \edit{modeling} human behavior can improve team performance across many applications, such as autonomous driving \cite{sadigh2016planning}, assistive robotics \cite{jeon2020shared}, and collaborative games \cite{pellegrinelli2016human}. Both model-free and model-based approaches have been employed for modeling human behavior. Model-free approaches \edit{(e.g., imitation learning \cite{carroll2019utility})} require substantial data and generally employ neural networks to learn human behavior without making strong assumptions.

In contrast, model-based approaches require far fewer samples but make certain assumptions about human behavior (e.g., humans exhibit bounded rationality \cite{simon1972theories}). Prior works in HRI have used POMDPs and their variants (e.g., BAMDP, MOMDP, I-POMDP) to account for latent factors such as trust, intent, or capability influencing human decision-making \cite{chen2018planning, lee2020getting, ramachandran2019personalized, wang2016impact}. Most prior POMDP-based works either assume known model parameters or employ maximum likelihood estimation (MLE) to estimate them \cite{lee2019bayesian, ramachandran2019personalized, chen2018planning, wang2016impact}. However, these approaches can fail to generalize to a diverse population and are prone to overfit \cite{bishop2006pattern}. Hence, we instead adopt a Bayesian approach to jointly learn the POMDP parameters and the robot policy during human interactions, similar to prior work \cite{lee2020getting, nanavati2021modeling, ng2012bayes}. However, a major drawback of such Bayesian approaches is the need to update beliefs over an augmented state space comprising both the human latent states and the POMDP parameters, which can quickly become computationally intractable. We overcome this challenge by making key approximations about the belief space and using conjugate priors for belief representation, which allows for computing quick belief updates. Our work differs from prior \old{Bayesian approaches in HRI} \cite{lee2020getting, nanavati2021modeling} \old{by maintaining} belief about \emph{dynamic} latent parameters, such as trust or compliance \cite{natarajan2020effects, chen2018planning} which varies during interactions and across individuals. 
\vspace{-2pt}
\subsection{Human-Agent Teaming}
Recently, there has been a surge in interest in designing AI agents that are capable of collaborating with humans, especially in ad-hoc settings \cite{barrett2017making, hu2020other, hong2023learning}. Ad-hoc or zero-shot human-agent teaming requires agents to be adept at collaborating with diverse users in novel contexts without prior interactions. Achieving ad-hoc, zero-shot coordination with novel human partners has been a longstanding challenge in AI and will be crucial for the ubiquitous deployment of robots and AI agents \cite{klien2004ten, paleja2021utility}. Recent works aim to achieve ad-hoc human-AI teaming either from human-human demonstrations using Behavior Cloning \cite{carroll2019utility} and offline RL \cite{hong2023learning} or via self-play without any human data \cite{strouse2021collaborating}. Others have also explored population-based training to learn robot policies that are generalizable across diverse users \cite{zhao2023maximum, lou2023pecan}. However, these approaches focus on domains where both humans and agents have symmetric capabilities and work concurrently. In contrast, our work examines mixed-initiative teaming, where humans and agents possess \emph{asymmetric capabilities} and must share control to achieve the task objective. Thus, we cannot learn robot policies from human-human demonstrations. Further, we seek to optimize team performance when all teammates are \emph{suboptimal}, which is seldom explored in human-robot teams \cite{lee2020getting, natarajan2023human}. 

\section{Preliminaries}
We model the human-robot team as a Bayes-Adaptive POMDP (BA-POMDP)  \cite{ross2007bayes}, allowing the robot to dynamically learn and adjust its policy based on estimations of human model parameters, while accounting for estimation uncertainty. 

A POMDP is defined as a tuple $\mathcal{M} = (S, A, O, \mathcal{T, E, }\: d_0, R, \gamma)$ where $S$ is a set of states $s \in S$, $A$ is a set of actions $a \in A$, 
$O$ is a set of observations $o \in O$, $\mathcal{T}(s_{t+1}|s_{t}, a_{t})$ is the state transition probabilities, $\mathcal{E}(o_t|s_t)$ is the emission function, $d_0$ is the initial state distribution, $R(s_t, a_t)$ is the reward for taking action $a$ in state $s$ at time step $t$, and $\gamma \in (0, 1]$ is the discount factor. \edit{The agent's goal is to learn a policy, $\pi: \mathcal{B} \rightarrow A$, that maximizes the expected cumulative discounted reward (return), \old{where $b \in \mathcal{B}$ is a belief state inferred by a history of previous observations and actions, $h$}.}
Belief updates can be achieved via the Bayes rule (infeasible for large state spaces) or with an unweighted particle filter (approximate update). 


Most prior works in POMDPs assume a fully-specified environment 
(i.e., the model parameters $\mathcal{T}$, $\mathcal{E}$ are known) \cite{lauri2022partially}, which is unrealistic in HRI as we neither have access to the person's true latent states (e.g., trust, preferences) nor how they change during the interaction. 
We adopt the BA-POMDP framework --- a Bayesian Reinforcement Learning approach for solving POMDPs \cite{ross2007bayes}. The BA-POMDP employs Dirichlet vectors, $\chi$, to represent uncertainty over the model parameters $(\mathcal{T}, \mathcal{E})$. As the POMDP states are hidden, $\chi$ cannot be computed and is included as part of the state.

\subsection{Solving POMDPs}
Partially Observable Monte-Carlo Planning (POMCP) is an online solver that extends the Monte-Carlo Tree Search (MCTS) to POMDPs \cite{silver2010monte}. 
POMCP uses the UCT (Upper Confidence Bound (UCB) for Trees) to select actions and an unweighted particle filter for belief updates. In POMCP, the UCT algorithm is extended to partially observable domains using a search tree of histories $h$ instead of states, where each node in the tree stores statistics -- visitation count $N(h)$, value or mean return $V(h)$, and belief $b(h)$, approximated by particles. The algorithm performs online planning through multiple simulations, incrementally building the search tree.
The return of each simulation is used to update the statistics for all visited nodes. 
POMCP terminates based on preset criteria (e.g., maximum number of simulations).

We model the human-robot team as a BA-POMDP. Solving BA-POMDPs is difficult as they are infinite-state POMDPs. 
\edit{The current state-of-the-art, online algorithm for solving BA-POMDPs is BA-POMCP}(extending POMCP for BA-POMDPs)~\cite{katt2017learning}. 
\edit{In this work, we propose Bayes-POMCP, which extends the BA-POMCP algorithm for suboptimal human-robot teams}.


%

\section{Method}
In this section, we first define the human-robot team model \old{(BA-POMDP)} for mixed-initiative interactions and then describe how we utilize a variant of the BA-POMCP algorithm to learn an adaptive robot policy for our current setting.

\subsection{Human-Robot Team Model} 

\subsubsection{State Space} 
In our human-robot team model, the state space combines the world state and user latent states $s = (x, z)$. The world state, $x \in \mathcal{X}$, refers to the task that the human-robot team is working on, and the latent states, $z \in \mathcal{Z}$, can refer to the user's trust or tendency to comply with the robot and their task execution preferences.
The robot does not have access to the user's latent states and must infer these states by observing the user's actions.
We focus on suboptimal human-robot teaming, assuming that the suboptimality arises from task-related errors or incomplete knowledge, i.e., both agents may make errors or cannot observe the full world state. Thus, the world state as observed by the robot may not always align with what the human observes $(x^R_t \neq x^H_t, \forall t)$. 

\subsubsection{Action Space} 
As we are planning from the robot's perspective, the action space comprises the actions $a^R \in A^R$ that the robot can take in the environment. In our mixed-initiative collaborative scenario, we assume that the robot first observes the human action and then selects its action\footnote{\old{Our approach is not restricted to this mixed-initiative setting and can be extended to cases where either the robot takes the first action or works concurrently with users.}}. The robot can choose to either execute, intervene, or override the user's actions. Additionally, the robot may choose to explain whenever it intervenes or overrides the user.

\subsubsection{Observation Space} 
The robot observes the human actions $a^H \in A^H$. 
We assume that the human's action depends on their knowledge of the current world state $x_t$ and the history of interactions, $h_{t-1}$ with the robot, i.e., the human follows the policy, $\pi^H(a^H_{t}|x_{t}, h_{t-1}, a^R_{t-1})$, where $h_{t-1} = \{a^H_0, a^R_0, a^H_1, a^R_1, \cdots, a^H_{t-1}\}$. Similar to prior work \cite{chen2018planning}, we assume that the user's latent state, $z_t$, is a compact representation of the interaction history ($z_{t} \approx \{h_{t-1} \cup a^R_{t-1}\}$). Thus, $\pi^H(a^H_{t}|x_{t}, h_{t-1}, a^R_{t-1}) \approx \pi^H(a^H_{t}|x_{t}, z_t)$.

\subsubsection{Transition and Emission Models} We define the state transition model, $\mathcal{T}$, from the robot's perspective, i.e., $\mathcal{T} = p(s_{t+1}|s_t, a^R_t)$. 
However, for mixed-initiative settings, the transitions in the state, $s_t = (x_t, z_t)$, occur as a result of both human and robot actions at each time step. Thus we rewrite the transition model as:

\begin{eqnarray}
p(s_{t+1}|s_t, a^R_t) = \sum\limits_{a^H} p(s_{t+1}|s_t, a^R_t, a^H_t) \times \pi^H(a^H_t|x_t, z_t) \\
= \sum\limits_{a^H} p(x_{t+1}|x_t, a^R_t, a^H_t) \times p(z_{t+1}| z_t, a^R_t, a^H_t) \times \pi^H(a^H_t|x_t, z_t)
\label{eqn:independence}
\end{eqnarray}
Equation~\ref{eqn:independence} comes from our assumption that given the human \old{and robot }actions, the world state dynamics are independent of the human latent state dynamics. In our collaborative scenario, we only estimate the latent state dynamics as part of the BA-POMDP, as we assume that the world state dynamics are deterministic and known.

The emission model $\mathcal{E}$ for the human-robot team refers to the human policy $\pi^H(a^H_{t}|x_{t}, z_t)$ which is also unknown to the robot and must be estimated to solve the BA-POMDP.

\subsubsection{Reward Function} 
\old{The reward function $\mathcal{R}(x, a^H, a^R)$ is positive for team actions that contribute to achieving the task goal and negative for team actions that hinder task success. We assume that both the user and the robot are aware of the reward function.}

\subsection{\edit{Adaptive Robot Intervention Policy in Mixed-Initiative Teams (Bayes-POMCP)}}
To maximize human-robot team performance in \old{real-time for} mixed-initiative settings, we implement a modified version of the BA-POMCP~\cite{katt2017learning}. 
Here, we highlight the key changes we make to the BA-POMCP algorithm. Figure~\ref{fig:overview} provides an overview of our approach, and the complete procedure is described in Algorithm~\ref{algo:BayMax}.

\begin{figure*}[t]
    \centering
    \includegraphics[width=.85\textwidth]{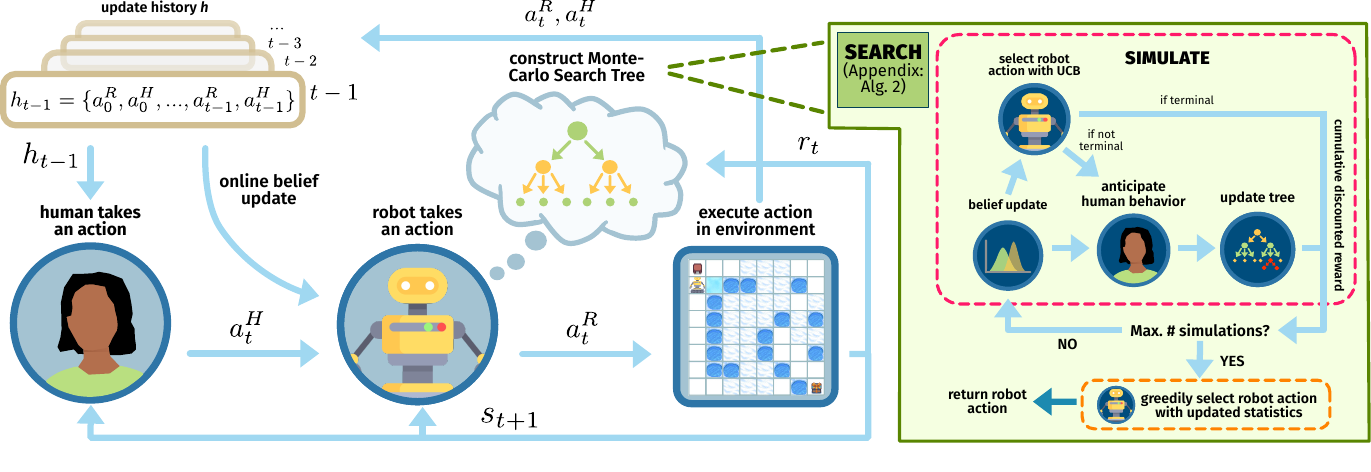}
    \caption{Graphical overview of the Bayes-POMCP approach for mixed-initiative Human-Robot Teaming: At each timestep $t$, the human first takes an action based on interaction history, $h$, and their current observation of the world state, $x$. The robot then determines when and how to intervene by anticipating human behavior using a Monte-Carlo tree search. The reward is calculated based on both human and robot actions.}
    \label{fig:overview}
    \label{fig:overview}
\end{figure*}

\begin{algorithm}

\DontPrintSemicolon
\caption{\bf \model: Maximizing Performance in Mixed-Initiative Human-Robot Teams}
\label{algo:BayMax}
\KwIn{Initial world state $x_0$; Interaction history $h_{-1} = []$; initial belief $b_0$; Search Tree $T = \{\}$}
$a^R_{-1} \leftarrow \text{No-Assist}$ \codeComment{// By default before episode starts}\;
$z_{0} \leftarrow \{h_{-1} \cup a^R_{-1} \}$ \codeComment{// Initial human latent state}\;
$a^H_0 \leftarrow \textsc{RealHuman}(\cdot|x_0, z_0)$ \codeComment{// First human action}\;
$h_0 \leftarrow [a^H_0]$\;
$T(h_0) \leftarrow \textsc{ConstructNode}(T, h_0)$ \codeComment{// Construct root node}\;

\For{$t = 0,1,2,\dots \text{max\_steps} \:$}{
    $\color{skyblue} a^R_t \leftarrow \textsc{Search}(h_t)$ \codeComment{// Root node $\rhd$ Search (Supp. Alg. 2)}\;
    \If{$(h^{}_ta^R_t) \notin T$}{
    \textsc{ConstructNode($T, h^{}_ta^R_t$)}\;
    }
    $x_{t+1} \leftarrow p(\cdot|x_t, a^R_t, a^H_t) \: \;$ \codeComment{// Update World State}\;
    $z_{t+1} \leftarrow \{h_t \cup a^R_t\} \: \; $ \codeComment{// Update true latent state $\not{\Rightarrow}$ Robot} \;
    $a^H_{t+1} \leftarrow \textsc{RealHuman}(\cdot|x_{t+1}, z_{t+1})$ \codeComment{// Next user action}\;
    $h_{t+1} \leftarrow h_t \cup \{a^R_{t}, a^H_{t+1}\}$\;
    \If{$h_{t+1} \notin T$}{
        $T(h_{t+1}) \leftarrow $  \textsc{ConstructNode($T, h_{t+1}$)}\;
    }
    \codeComment{// Belief update: next root node}\;
    $\color{skyblue} b(h_{t+1}) \leftarrow \textsc{Belief-Update}(b(h_t), a^R_t, a^H_{t+1})$
    
    $\textsc{Prune-Tree}(T, h_{t+1})$ \codeComment{// $h_{t+1}$ is the root node} \;
    
}

\end{algorithm}

\subsubsection{Belief Approximation} Similar to POMCP, BA-POMCP is an online algorithm that constructs a lookahead search tree through environment simulations and maintains a belief over latent parameters using an unweighted particle filter to determine the best action at each time step. However, in BA-POMCP, we need to maintain a belief over both the latent states $|S|$ and the model parameters $\mathcal{T}, \mathcal{E}$ ($ |S|^2 \times |A| + |S| \times |A| \times |O|$ parameters). Computing the posterior update over such a large space can be expensive. Further, it is difficult for the posterior distribution to converge to the true parameters, especially when we only have access to limited interactions. 

Hence, we leverage the independence assumption between the world state and the latent state transition (Equation~\ref{eqn:independence}) to approximate the belief in each node in the search tree. Approximating the belief makes it feasible to compute the belief updates in real-time for fluent HRI. Since we only need the human action to determine the next world state, we choose to maintain the belief only over the \old{user policy} space instead of all latent states and model parameters. We compute the posterior update for the belief \old{$b(h_{t+1})$ from the prior belief, $b(h_t)$, based on the interaction history, $h_t$, at each node}.

\subsubsection{Simulating Human Policy} In BA-POMCP, we need to simulate human actions during the rollout for constructing the search tree. 
As the robot lacks direct knowledge of the true human policy, we first estimate the human policy parameters and use the same for simulation.
Given that the human actions can be categorized as being compliant/non-compliant with the robot,
we model the true human policy as a Bernoulli distribution with an unknown parameter, $\mu$, that signifies the likelihood of user compliance for a given interaction history, $h$. \old{To estimate $\mu$, we adopt a Bayesian approach. We assume a prior distribution or belief over the space of human policies $b = p(\mu)$.} We approximate $b$ using \old{a set of particles, which is updated upon subsequent interactions with the user. In general, performing the belief update can be computationally expensive, but such updates can be computed efficiently for the conjugate family of distributions \cite{bishop2006pattern}. Thus, we} model each particle as a beta distribution -- the conjugate prior for Bernoulli distributions. 

To simulate the human action during rollout, we sample a particle from $b$ at the current node. We use this \edit{sampled} particle to anticipate \edit{}the next human actions and update it based on the interaction outcomes during the simulation. \edit{Additionally, we assume that humans are rational and employ an $\epsilon$-greedy heuristic to select the user's actions in case of non-compliance.}

Alternatively, we can use a random policy to mimic human behavior, but this would require more simulations to cover a range of possible human responses and determine the optimal robot action---resulting in increased computation time. Therefore, we opt for estimating \edit{user compliance} and then simulating the human policy, which we find empirically more efficient.

To evaluate the contributions of our proposed modifications to the BA-POMCP algorithm \cite{katt2017learning}, we perform an ablation analysis without modeling humans, i.e., we only use random rollout policies for anticipating human behavior and perform no belief updates. We refer to this approach as POMCP in our analysis (Section~\ref{sec:sim_experiment}). 


\section{Evaluation}
\subsection{Domain} We modified the Frozen Lake environment from OpenAI Gym \cite{brockman2016openai} for \old{evaluating} mixed-initiative human-robot teaming. In this domain, the users must collaborate with the robot to navigate a\old{n $8 \times 8$} frozen lake grid from start to goal in the fewest steps possible while avoiding holes and slippery regions. We modified the original domain to only have certain grids as slippery instead of a constant slip probability throughout the map. Stepping on a slippery region will cause the agent to fall into a hole. Both the human and the robot can only observe whether the \old{adjacent four grids are} slippery. \old{Each time the agent falls into a hole, the team incurs a penalty $\alpha$ and must begin again from the start location.} 

\begin{figure}[t]
    \centering
    \includegraphics[width=\columnwidth]{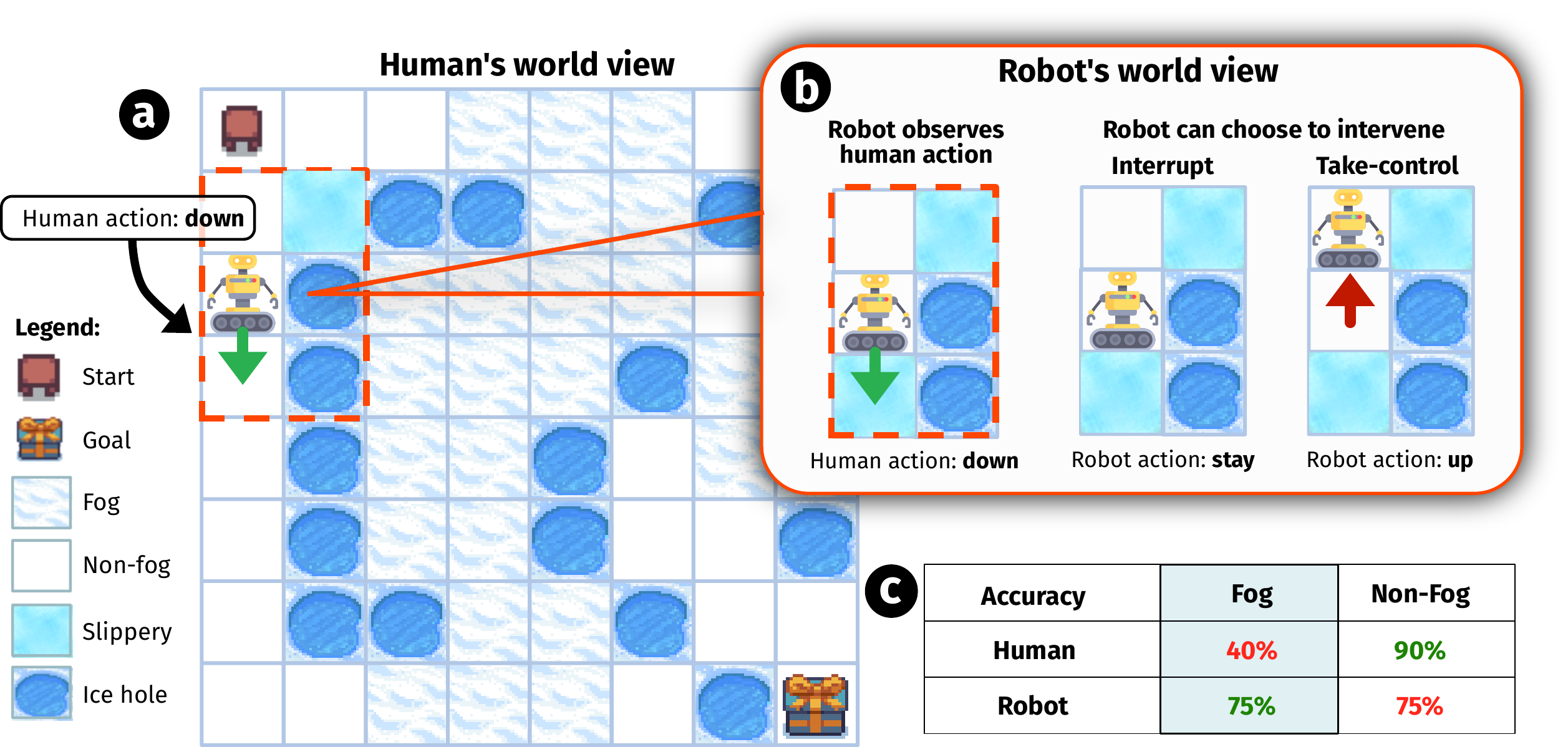}
    \caption{Frozen Lake Domain used in this study. Figure 2(a) shows the overall game layout. Figure 2(b) depicts robot intervention styles: interrupt, take-control, and Figure 2(c) shows the human and robot accuracies in identifying slippery grids.}
    \label{fig:setup}
\end{figure}

To enforce suboptimality, we introduce errors in the human and robot observations of slippery grids. These errors include -- \textbf{False Positives} (observing a safe grid as slippery), and \textbf{False Negatives} (observing a slippery region as safe). Moreover, certain parts of the map are covered by fog which reduces human visibility. The human and robot accuracies for identifying slippery regions are shown in Figure~\ref{fig:setup}. During the game, the human teleoperates the robot across the lake, but the robot may intervene or take control if it finds that the user chose a longer or unsafe path (e.g., slippery regions or holes) to the goal. 
Additionally, the user is equipped with a high-quality ($100\%$ accurate) sensor for detecting slippery regions in adjacent grids, but each use of the sensor incurs a point cost \old{$\rho$. The overall team performance or game reward for each round is calculated as a combination of step penalty (shorter path $\rightarrow$ higher reward), penalty for falling into holes $\alpha$, detection penalty $\rho$, and a bonus $\kappa$ for reaching the goal} as shown in Equation~\ref{eqn:reward}.
\begin{multline}
    \label{eqn:reward}
    \text{Reward} = \textrm{Max steps} -\textrm{\# steps taken} -\alpha \times \textrm{\# falls into hole} \\
    -\rho \times \textrm{\# detections} + \kappa \times \mathbbm{1}[\textrm{goal reached} == \textrm{True}]
\end{multline}

\old{We empirically set max steps $=80$, $\alpha=10$, $\rho=2$ and $\kappa=30$ for our human-subject experiments.} Our environment is inspired by USAR missions, where humans teleoperate robots, but both humans and robots can have complementary skills and varying domain knowledge. Further details of the user study domain can be found in the \textcolor{black}{Supplementary}.

\subsection{Human-Subjects Experiments} 
We conducted two user studies
to 1) examine how users respond to different robot intervention styles with and without explanations but with a static policy
(\textbf{Data Collection Study}) and 2) evaluate human-robot team performance with the proposed adaptive \old{Bayes-POMCP} approach (\textbf{Evaluation Study}). 

\subsubsection{Data Collection Study} We employ a $1 \times 5$ within-subjects experiment design to examine user responses to various robot interventions \old{in mixed-initiative teaming (Figure~\ref{fig:setup}b)}. These interventions include -- \emph{no assist}: the robot does not intervene (baseline), \emph{interrupt}: the robot stops the user from executing an action, \emph{take-control}: the robot overrides the user's action with its own action, \emph{interrupt+explain}: the robot interrupts and explains, \emph{take-control+explain}: the robot takes over control and explains. 
To ensure consistency across intervention strategies, the robot employs the same handcrafted heuristic that determines when to intervene. 
The heuristic intervention policy \old{is a short-horizon planner that only intervenes if the user's current action is anticipated to lead to a slippery region (based on the robot's knowledge), a hole, or a longer path ($\geq $k steps) and will cede control to the user if the user persistently chooses the action the robot is intervening.
The heuristic employs a static intervention style. The algorithm for the heuristic policy can be found} in the \textcolor{black}{Supplementary}.

\begin{figure*}
     \centering
     \begin{subfigure}[b]{0.44\textwidth}
         \centering      \includegraphics[height=0.2\textheight]{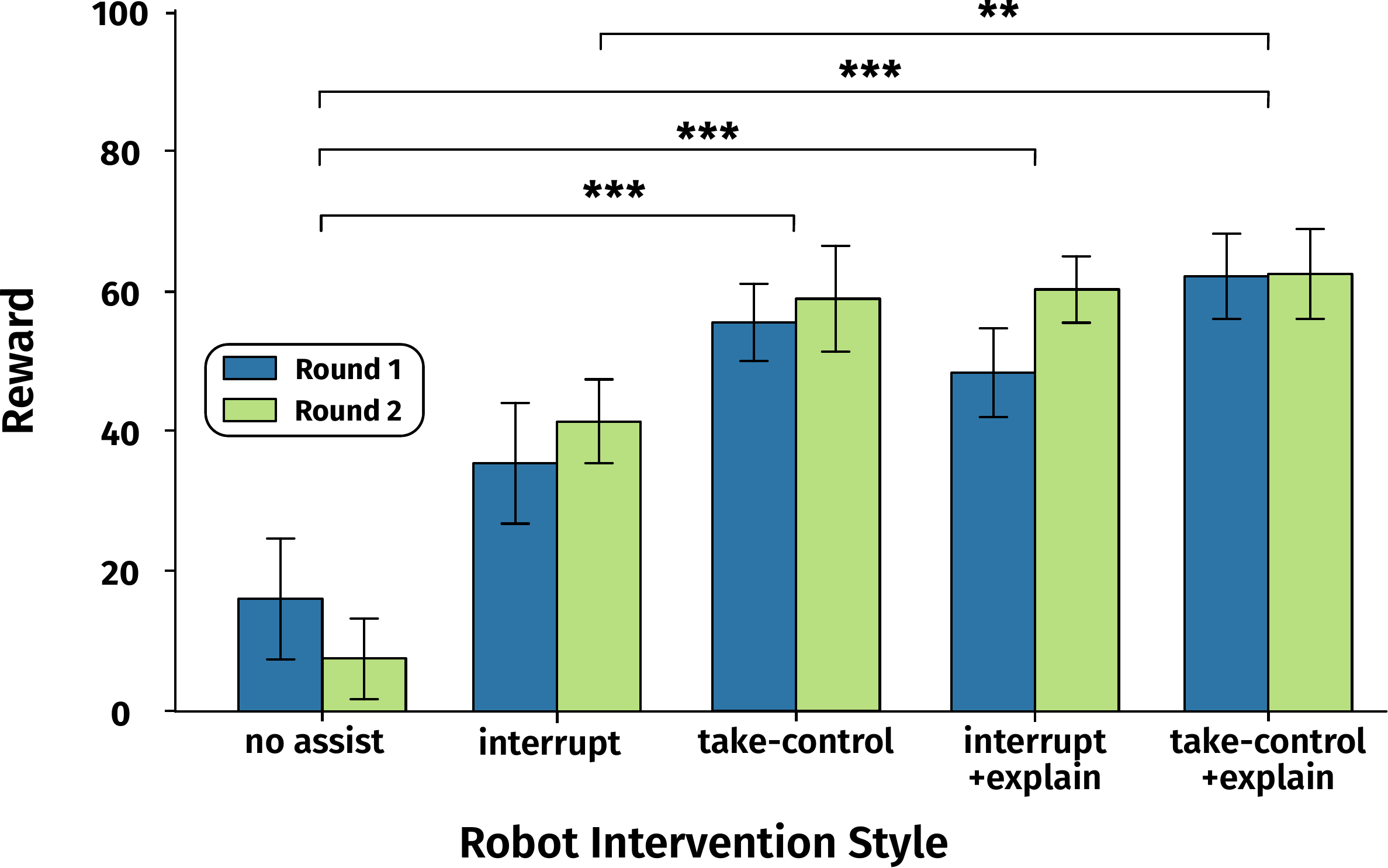}
         \caption{Team Performance vs. Robot Intervention Style}
         \label{fig:score_1}
     \end{subfigure}
     \hfill
     \begin{subfigure}[b]{0.44\textwidth}
         \centering
         \includegraphics[height=0.2\textheight]{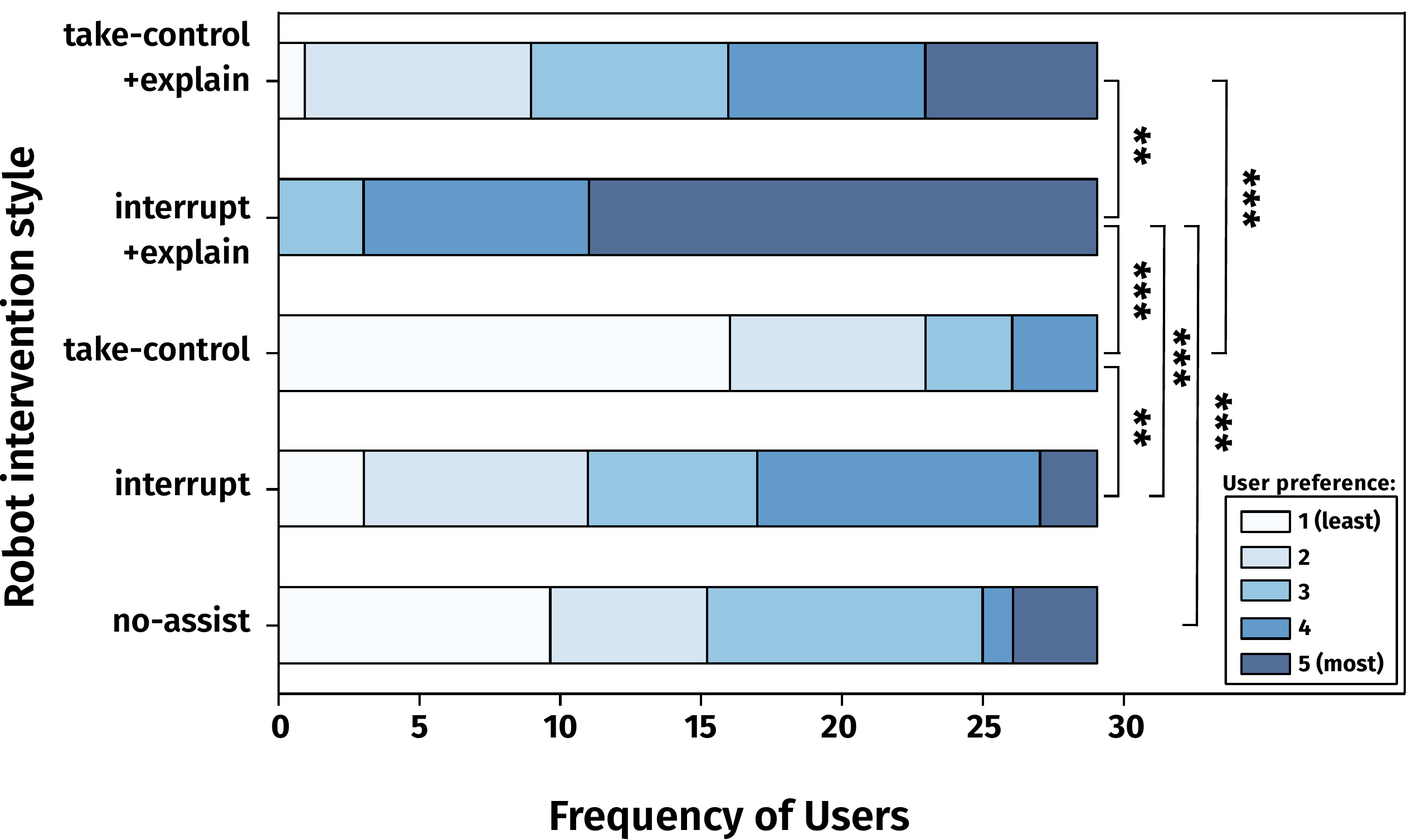}
         \caption{User preferences for working with different robots}
         \label{fig:pref_1}
     \end{subfigure}
     \hfill
        \caption{Results from Data Collection Study \old{with Heuristic Mixed-Initiative Policies}. Figure~ 
 \ref{fig:score_1} shows that the team performance is the highest for the \emph{take-control} agents and the lowest with no-assist (baseline). Figure~\ref{fig:pref_1} shows the users preference ranking across intervention styles. The majority of the users prefer to work with the \emph{interrupt+explain} agent the most (rank $= 5$).}
        \label{fig:study_1}
\end{figure*}

\subsubsection{Evaluation Study}
We employ a $1 \times 3$ within-subjects experiment to compare human-robot team performance under different robot policies. The examined policies are our
proposed approach -- \old{Bayes-POMCP}, the same heuristic policy as was used in the data collection study, and an adversarial policy (\old{Adv-Bayes-POMCP}) optimized for \old{negative} game reward (Equation \ref{eqn:reward}). 
We include the adversarial policy as an adaptive baseline to show that (1) our proposed approach can successfully aid or inhibit the user from reaching the goal, and (2) it is essential for the adaptive policy to reason when to intervene effectively in addition to switching the intervention styles.
\old{To perform a balanced comparison, we ensure that the run times of all robot policies are identical. }
Further, we limit the use of the detection sensor ($\leq 5$) in the evaluation study to force participants to rely on the robot's assistance.

\subsubsection{Metrics}
For both studies, we assess user preferences and performance using subjective and objective measures, respectively. Our subjective measures include trust \cite{muir1989operators}, likeability \cite{bartneck2009measurement}, and willingness to comply \cite{raemdonck2013feedback} (adapted from human-human interactions for HRI) measured via 5-point Likert scales. 
All questionnaires were administered to the users after each round in both studies. 
Further, participants reported their demographics, highest completed education, prior experience with robots, and completed a 50-item personality scale \cite{Goldberg1992TheStructure} as part of the pre-study questionnaire. At the end of the study, users ranked their preferences for the different robot agents. All questionnaires used for the study can be found in the \textcolor{black}{Supplementary}. Objective performance was assessed based on the total game reward (Equation~\ref{eqn:reward}) in each round.


\subsubsection{Participants and Procedure}
 We recruited $30$ participants (Age: $25.56 \pm 3.38$, Female: $33\%$) for the data collection study and $28$ new participants (Age: $25.27 \pm 3.28$, Female: $50\%$) for the evaluation study, all from a local university campus after IRB approval. 
 The procedure was the same for both studies. 
 Written consent from the participants was obtained before the experiment.
 At the start of the study, participants received written game instructions along with a demonstration from the experimenter. 
Participants first completed three practice rounds to familiarize themselves with the game and then engaged in ten and six rounds (two rounds per condition) for the data collection study and evaluation study, respectively. The subjects were instructed to complete each round by taking the shortest path to the goal. 
The experiment order was randomized, and participants completed pre- and post-study questionnaires.

\subsection{Hypotheses}
To investigate how different users interact with various robot intervention styles, we first conducted a data collection study. We hypothesize the following:

\vspace{4mm}
\noindent \textbf{H1A: }\emph{The human-robot team performance can vary with different robot intervention styles.} Although the robot follows the same heuristic across different conditions in Study 1, we hypothesize that the team performance will vary across intervention styles as users may respond differently. For instance, users may be better informed to choose the next action appropriately when the robot intervenes and provides an explanation.

\vspace{4mm}
\noindent \textbf{H1B: }\emph{Users will have different preferences for working with various robot intervention styles.} Humans have varying personality traits and task preferences, which may impact how they perceive and collaborate with teammates. For instance, extroverted individuals are more likely to assume leadership and less likely to renounce control in human-human teams \cite{kickul2000emergent}. Likewise, we hypothesize that users will have different preferences when working with robots that interrupt or take control with or without offering explanations.

\vspace{5mm}
\noindent For the evaluation study, we compare the human-robot team performance with the adaptive \old{Bayes-POMCP} policy against heuristics used in the first study and an adversarial baseline -- \old{Adv-Bayes-POMCP}. We hypothesize:

\vspace{4mm}
\noindent \textbf{H2A: }\emph{The human-robot team performance will be the highest when the robot employs the adaptive \old{Bayes-POMCP} policy.} We hypothesize that the \old{Bayes-POMCP} policy which actively anticipates human actions by considering their latent states, is better suited for determining when and how to intervene various users and will thereby maximize team performance. In contrast, the baselines that do not model the human latent states (the heuristic policy) or optimize for negative reward (the adversarial Bayes-POMCP), will not be able to assist the users appropriately.

\vspace{4mm}
\noindent \textbf{H2B: }\emph{Users will most prefer to work with our proposed approach, the adaptive \old{Bayes-POMCP} policy.} We hypothesize that the \old{Bayes-POMCP} policy can effectively intervene users by modeling their latent states and will, therefore, not only improve team performance but also have a positive impact on the users' subjective preference for collaborating with the robot.

\section{Results and Discussion}
\label{sec:results}
In this section, we first discuss the results of the data collection study. Next, we show results from our simulation experiments used to validate \old{Bayes-POMCP} before testing on human participants. We then discuss the results from the evaluation study, comparing our \model{} approach and two baselines.

All our statistical analyses were performed using libraries in R, and the significance level $\alpha$ was set at $0.05$. For our analysis, we use parametric tests unless the model fails to meet the required assumptions (normality, homoscedasticity, et cetera). Details of all models and tests used for each hypothesis, along with the effect sizes and statistical power, are listed in the \textcolor{black}{Supplementary}.

\subsection{Data Collection Study}
For the data collection study, we recruited $30$ participants and excluded one participant as an outlier since they failed to complete all ten rounds in the study (failure rate across all subjects: $1.733 \pm 2.365$). Thus we have data from $29$ subjects for our analysis.

\vspace{2mm}
\noindent \textbf{H1A: }\emph{Team Performance and Robot Intervention Styles.} We compare the team performance using the game reward (Equation~\ref{eqn:reward}) across the five robot intervention styles employed in the first study. The robot either used the same heuristic policy to determine when to intervene or did not intervene at all (no-assist: baseline condition). Each user participated in two rounds for each intervention style, totaling ten rounds, all played on different maps with varying levels of difficulty.  To mitigate ordering effects and map-related biases, the experiment conditions and map assignments were randomized.
\begin{figure*}
     \centering
     \begin{subfigure}[b]{0.32\textwidth}
         \centering
         \includegraphics[width=\textwidth, height=0.19\textheight]{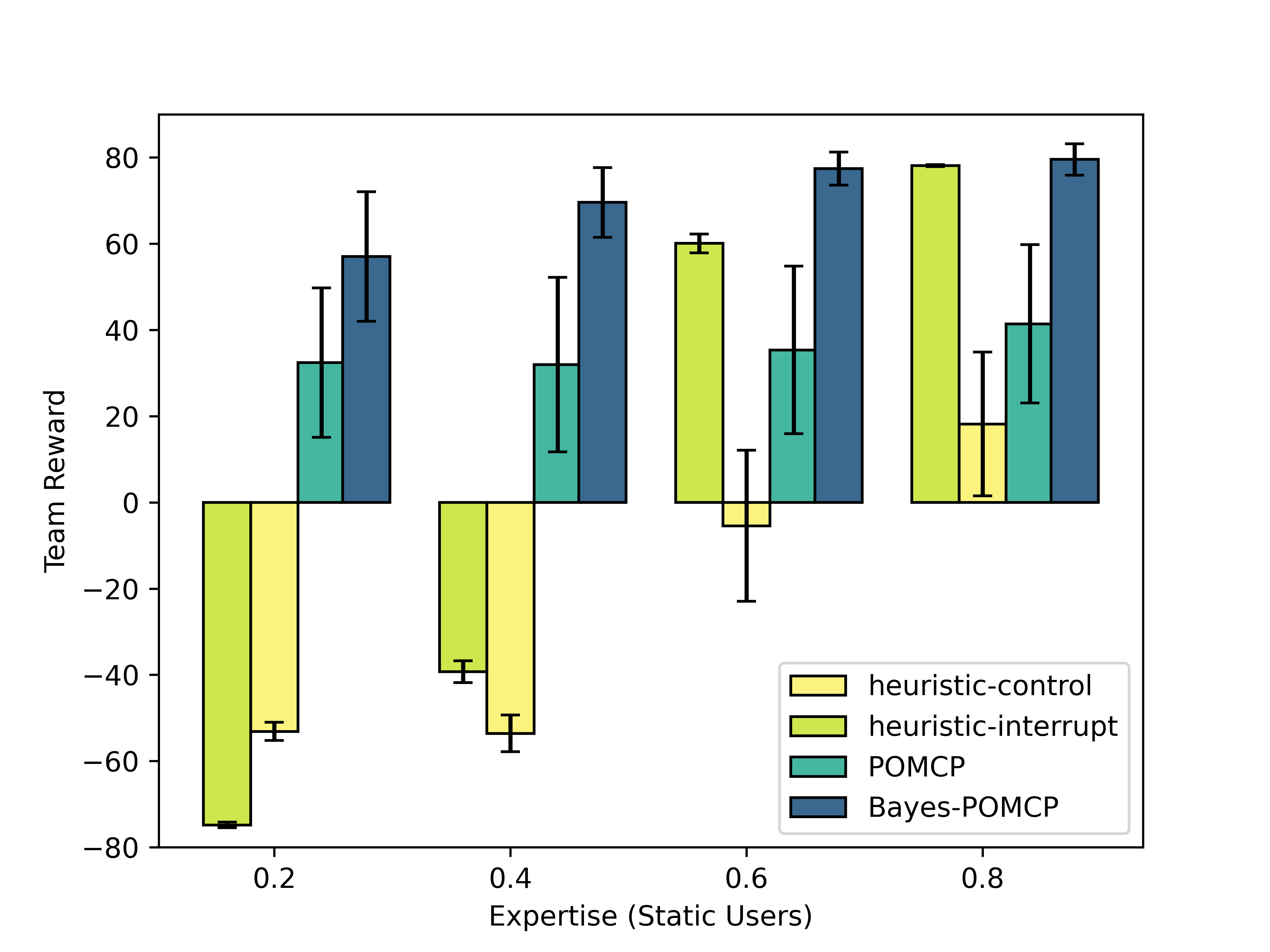}
         \caption{Static Users - vary expertise ($\psi$).}

         \label{fig:static_expertise}
                  \hfill
     \end{subfigure}
     \begin{subfigure}[b]{0.32\textwidth}
         \centering
         \includegraphics[width=\textwidth, height=0.19\textheight]{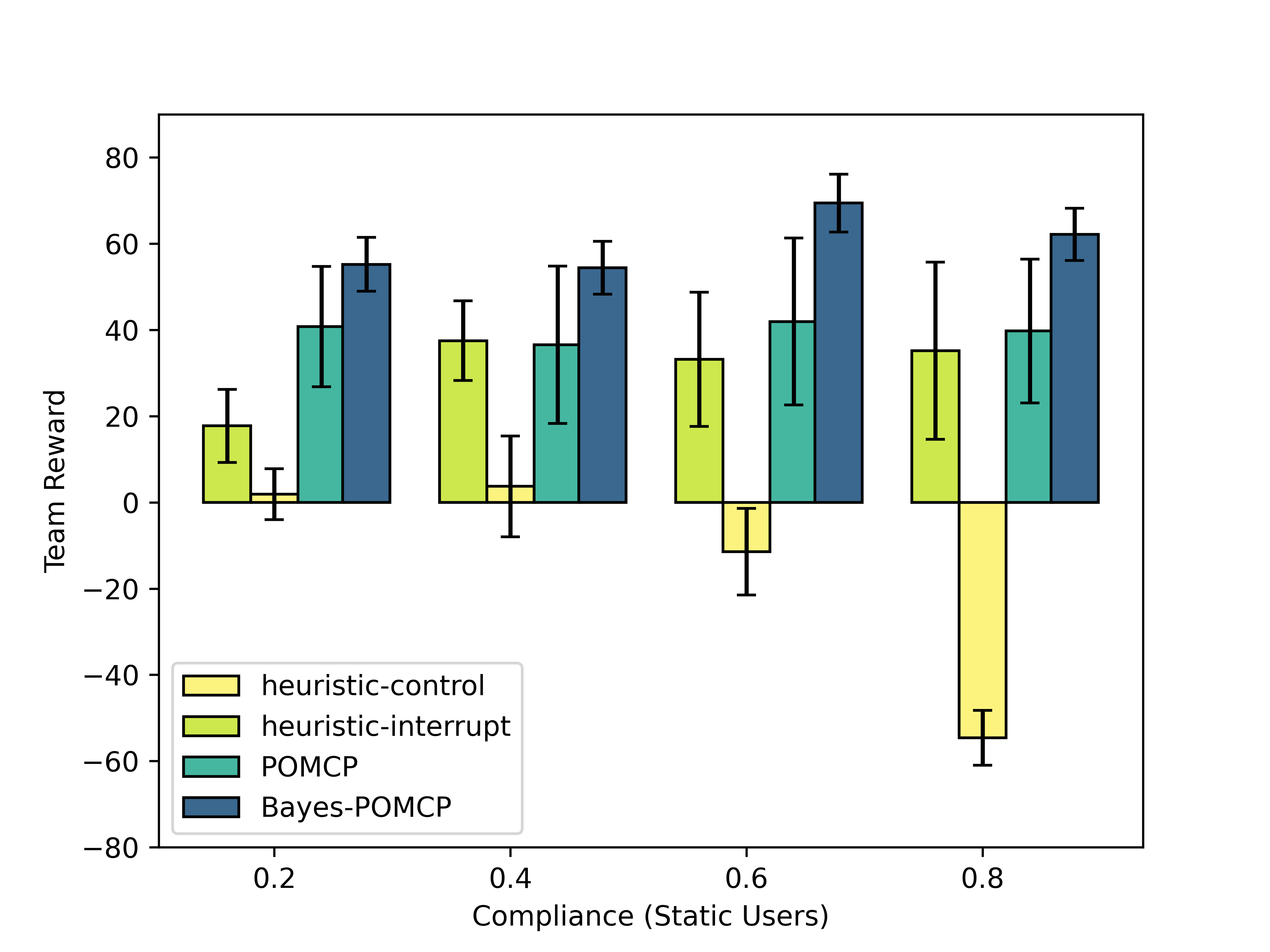}
          \caption{Static Users - vary compliance ($\theta$).}
    \label{fig:static_trust}
    \hfill
     \end{subfigure}
     \begin{subfigure}[b]{0.3\textwidth}
         \centering
        \includegraphics[width=\textwidth, height=0.17\textheight]{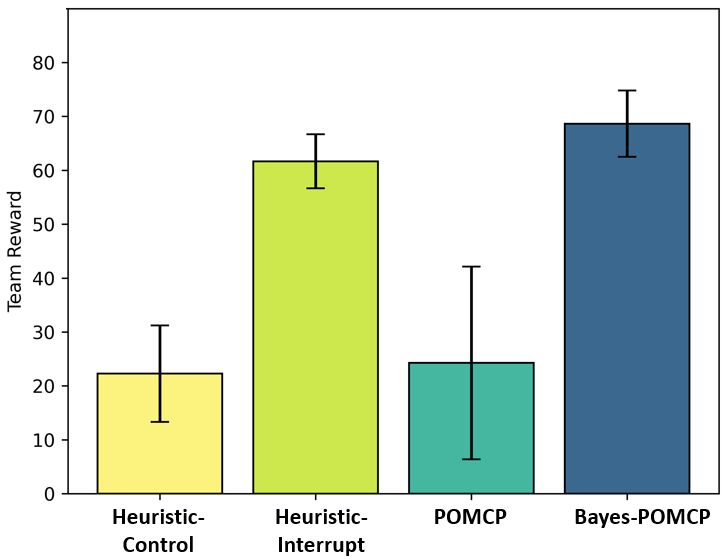}
         \caption{Dynamic Users}
         \label{fig:dynamic}
              \hfill
     \end{subfigure}
        \vspace{-4.5mm}
        \caption{Team performance in simulation experiments with static and dynamic latent user models. Figures~\ref{fig:static_expertise} and ~\ref{fig:static_trust} show that Bayes-POMCP can enhance team performance across users of varied expertise and compliance tendencies, respectively. Bayes-POMCP outperforms heuristics and the ablation POMCP model, especially for users with low expertise.}
        \label{fig:sim_exp}
\end{figure*}
We use Kruskal-Wallis (a non-parametric test), with the dependent variable as the reward and the independent variable as the robot intervention style. We obtain statistical significance for the intervention style ($H(4) = 58.16, p<.001$). Subsequently, we use Dunn's test for performing post-hoc pairwise comparisons, and the significance values are shown in Figure~\ref{fig:score_1}.

\vspace{2mm}
\noindent \textbf{Takeaway: }We find that the human-robot team performance is impacted by the intervention styles used by the robot, rejecting the null hypothesis (Figure~\ref{fig:score_1}).  Firstly, it is worth noting that the team performance significantly improves when the robot intervenes compared to the baseline (no assistance), validating the need for robot interventions in our study domain. Secondly, the team performance is the highest when the robot takes over control. Lastly, adding explanations did not significantly improve performance for the same intervention style (e.g., between interrupt and interrupt+explain).

\vspace{2mm}
\noindent \textbf{H1B: }\emph{Users' Working Preference and Robot Intervention Styles.} At the end of the first user study, participants were asked to rank their preferences for working with various robot intervention styles on a scale from 1 (lowest) to 5 (highest). As user rankings are considered as ordinal data, we use Kruskal-Wallis, a non-parametric test to analyze \textbf{H1B}. We find that robot intervention style indeed influences user preferences $(H(4)=61.67, p<.001)$. The majority of the users preferred the interrupt+explain agent the most and the take-control agent the least, as shown in Figure~\ref{fig:pref_1}.

\vspace{2mm}
\noindent \textbf{Takeaway: }Our results suggest that, despite explanations not improving performance, most users favor working with robots that offer explanations for their interventions. Interestingly, even though the take-control agent achieved the highest team performance, it was the least preferred choice for the majority of users. These findings highlight the need for an adaptive robot policy that adjusts the intervention style to maximize performance and user satisfaction. If the robot only takes over control, it can improve team performance in the short term but can cause user dissatisfaction and can potentially lead to users abandoning the system in the long run. 

\begin{figure*}
     \centering
     \begin{subfigure}[b]{0.45\textwidth}
         \centering      \includegraphics[height=0.2\textheight]{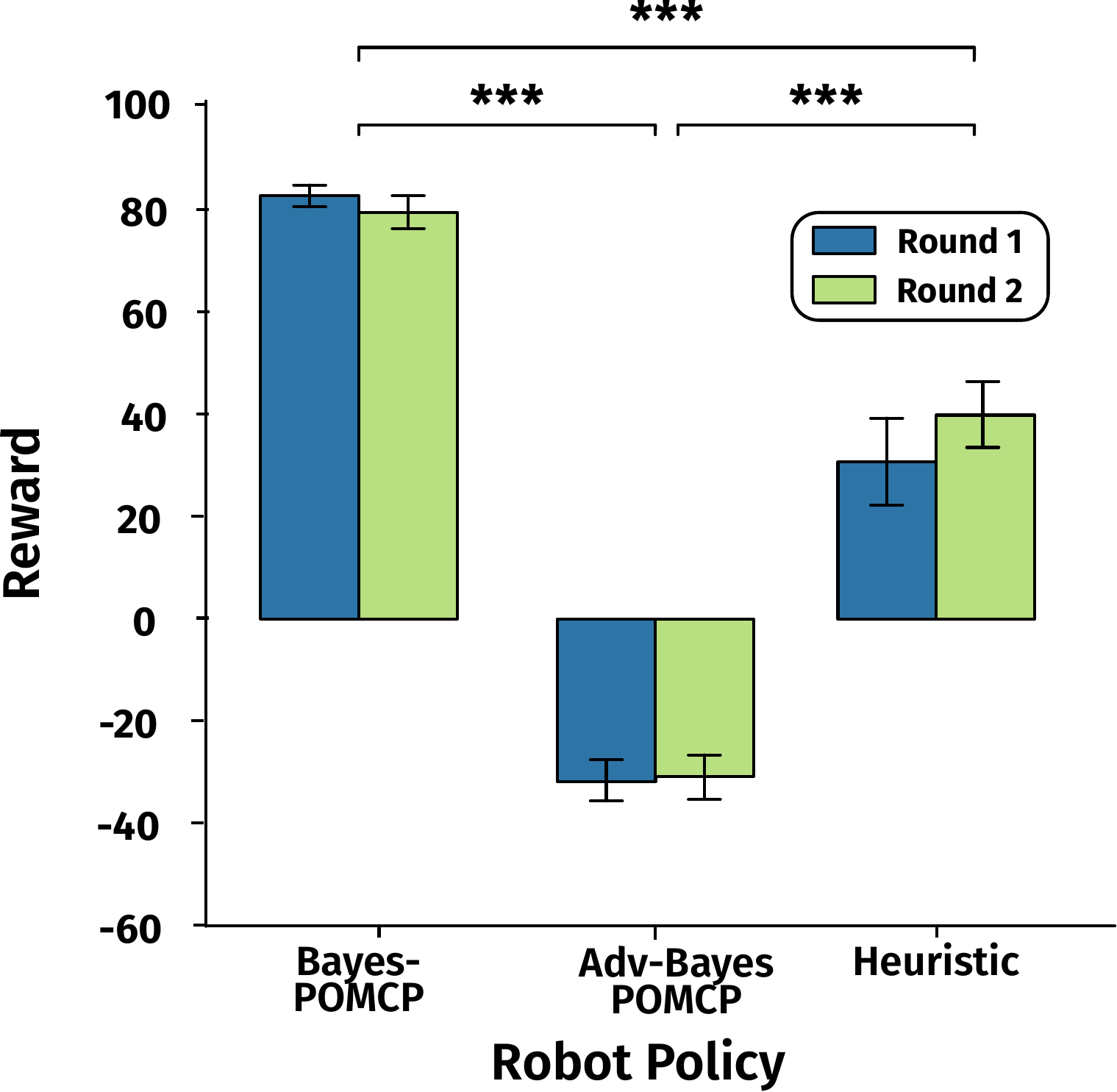}
         \caption{Team Performance vs. Robot Policies}
         \label{fig:score_2}
     \end{subfigure}
     \hfill
     \begin{subfigure}[b]{0.42\textwidth}
         \centering
         \includegraphics[height=0.2\textheight]{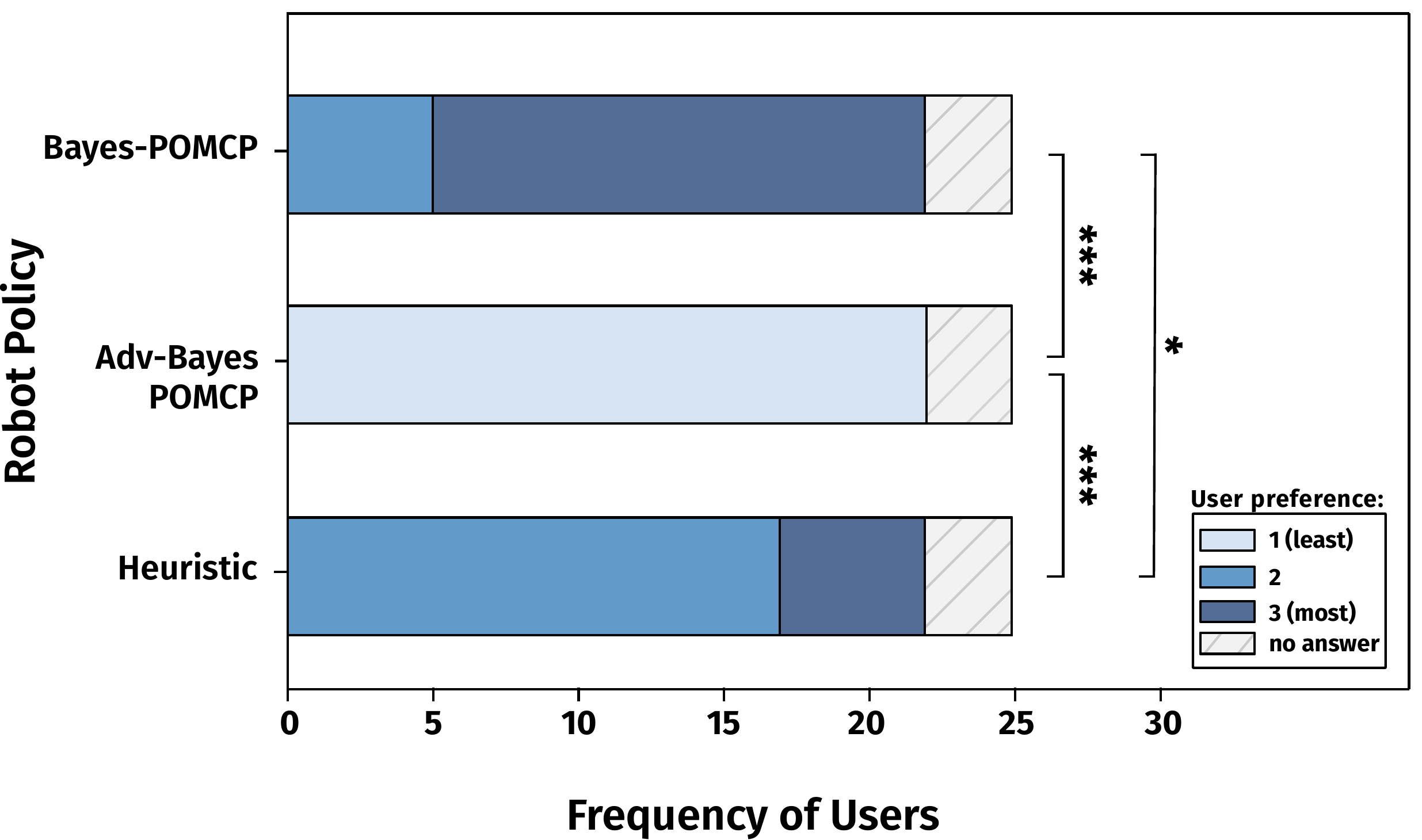}
         \caption{User Working Preferences for Robot Policies}
         \label{fig:pref_2}
     \end{subfigure}
     \hfill
        \caption{Results from the Evaluation Study. Figure~ 
 \ref{fig:score_2} shows that \old{the team performance is the highest for the }\emph{Bayes-POMCP} agent and the \old{lowest} for the Adv-Bayes-POMCP (the adversarial baseline). Figure~\ref{fig:pref_2} shows that the majority of the users prefer our approach compared to the baselines.}
        \label{fig:study_2}
\end{figure*}

\subsection{Simulation Experiments}
\label{sec:sim_experiment}
We first validate whether our proposed method can adapt to diverse users by testing with various simulated human models before testing the \old{Bayes-POMCP} policy on users. For the simulation experiments, we compare \old{Bayes-POMCP} against two baselines -- (1) the standard POMCP algorithm \cite{silver2010monte} with no human model \old{(POMCP)} and (2) the heuristic agents (both take-control and interrupt) on \old{five} of the $8 \times 8$ maps used in the data collection study. To simulate a diverse set of users, we modulate two latent parameters that determine their behavior -- the users' capability or expertise ($\psi$) to solve the task and the users' tendency to comply with the agent ($\theta$). We test with both static users (whose latent parameters -- $\psi, \theta$ are fixed) and dynamic users, whose $\theta$ varies continuously based on the interaction history, but $\psi$ remains fixed (i.e., we assume no learning effect as the domain is simple). We provide further details of the simulated human population in the \textcolor{black}{Supplementary}. Our results (Figure~\ref{fig:sim_exp}) indicate that Bayes-POMCP outperforms both the heuristics employed in the first study and the ablation baseline without human modeling (POMCP) for static and dynamic user models.

\subsection{Evaluation Study}
Upon verifying our policy with different simulated users, we collected data from $28$ new participants (who did not take part in the first user study) for the evaluation study. We excluded data from three subjects. Two of the three subjects encountered graphic rendering issues in the study interface. The other subject was excluded as an outlier as they failed to complete all six rounds (failure rate across all subjects: $3.48 \pm 0.77$). Hence we only include data from the remaining $25$ subjects for our analysis.

\vspace{2mm}
\noindent \textbf{H2A: }\emph{Team Performance and Robot Policy.} In this user study, we evaluated the team performance for different robot policies -- the heuristic agents (interrupt+explain and take-control+explain) from the first study, our proposed approach \old{Bayes-POMCP} optimized for the true reward and the negative reward. Each user participated in two rounds per policy, totaling six rounds, all played on different maps (a subset from the first study). We used the Kruskal-Wallis test with the reward as the dependent variable and the robot policy as the independent variable. We obtained statistical significance for the robot policy $(H(2) = 109.89, p<.001)$ and performed post-hoc analysis with Dunn's test, whose results are shown in Figure~\ref{fig:score_2}. 

\vspace{2mm}
\noindent \textbf{Takeaway: }We find that \old{Bayes-POMCP} policy significantly outperforms our baselines for team performance, rejecting the null hypothesis (Figure~\ref{fig:score_2}). We also find that the adversarial \old{Bayes-POMCP} is effective in preventing the user from reaching the goal, as reflected by the negative reward.  

\vspace{2mm}
\noindent \textbf{H2B: }\emph{Users' Working Preference and Robot Policy.} Users ranked their preferences for working with the different robot agents at the end of the second study. We perform the Kruskal-Wallis test, which shows that the robot policy significantly influences user preferences $(H(2)=45.41, p<.001)$. We find that $68\%$ of the users preferred the \old{Bayes-POMCP} agent the most, $20\%$ preferred \old{heuristic agents} the most, and $88\%$ preferred the Adv-\old{Bayes-POMCP} agent the least. $12\% \: (=3/25)$ did not answer the preference survey. 

\old{We also analyzed subjective metrics with Likert scales for trust, willingness to comply, and robot likeability. We conducted three rANOVA with the subjective metrics as the dependent variables and independent variables as robot policy, number of rounds completed, demographics (age, gender, prior robotics experience), and pre-study questionnaire responses of the user. We find that robot policy was statistically significant across all subjective metrics from the three ANOVAs, with our proposed approach having the highest mean values. We then performed post-hoc analysis using Tukey HSD. For further details of the analysis, see \textcolor{black}{Supplementary}.}

\vspace{2mm}
\noindent \textbf{Takeaway: }We find that \old{Bayes-POMCP} policy significantly outperforms our baselines across all subjective metrics, and the majority ($68\%$) of the users rated that they would most prefer to work with the Bayes-POMCP agent in the evaluation study.

\subsection{Summary of Results}
We summarize our key findings from two human-subject experiments and analysis with simulated human models: 
\begin{enumerate}
\item Robot interventions are necessary for improving team performance when both humans and robots are suboptimal due to having non-identical, partial domain knowledge.
    \item The robot intervention style (interrupt or take-control) can impact both team performance $(p < 0.001)$ and user preferences $(p < 0.001)$. Users prefer robots that offer explanations for interventions, albeit without performance improvement.
    \item Our proposed approach, Bayes-POMCP, can effectively intervene users (both simulated and real human subjects) to maximize human-robot team performance.
    \item Bayes-POMCP not only enhances team performance but also positively influences users' preference to collaborate with the robot and their self-reported measures, such as trust and likeability towards the robot.
    
\end{enumerate}

\section{Limitations and Future Work}
While our approach successfully improves human-robot team performance in a computationally efficient manner, 
Bayes-POMCP relies on an environment simulator to estimate the value of human-robot actions in the Monte Carlo search tree, which may not be available for real-world human-robot collaboration tasks. Therefore, in future work, we aim to explore alternative methods, such as deep learning \cite{silver2018general} for value estimation.
Moreover, our findings indicate that while robot explanations positively influenced users' subjective perceptions, they did not improve team performance. We hypothesize that this may be because the task was relatively simple, and users did not need explanations from the robot to enhance their decision-making. In future work, we seek to assess the utility of explanations in improving team performance for more complex teaming tasks. 
Finally, our findings are limited to short-horizon interactions, as users only played two rounds of the game with each agent. To address this limitation, our proposed approach can be extended to longitudinal HRI tasks, where robots must anticipate and adapt to changes in user behavior or preferences over time.

\section{Conclusion}
In this work, we propose an online Bayesian approach, Bayes-POMCP, to optimize performance in mixed-initiative human-robot teams when both agents are suboptimal.  Our focus is on learning a robot policy for effective user intervention. 
We find that robot interventions can improve performance while recognizing diverse user preferences. Next, we evaluate Bayes-POMCP, and show its effectiveness in improving team performance across different simulated human models and real users. We address the computational challenges in solving POMDPs by using a Monte-Carlo search with belief approximation and using conjugate priors to perform belief updates efficiently. In future work,
we plan to continue evaluating our algorithm for long-horizon interactions and extend it beyond grid-world domains to real-world human-robot collaboration tasks.

\begin{acks}
\old{This work was sponsored by a gift from Konica Minolta and the National Institutes of Health (NIH) under Grant 1RO1HL157457.}
\end{acks}

\bibliographystyle{plain}
\bibliography{references.bib}

\end{document}



\pagestyle{fancy}
\fancyhead{}


\maketitle

\begin{table}[]
\fontsize{7.5pt}{9.5pt}\selectfont
\centering
\begin{tabular}{|c|l|ll|ll|}
\hline
\multirow{2}{*}{Algorithm}   & \multicolumn{1}{c|}{\multirow{2}{*}{Map}} & \multicolumn{2}{c|}{100 simulations}                     & \multicolumn{2}{c|}{500 simulations}                     \\ \cline{3-6} 
                             & \multicolumn{1}{c|}{}                     & \multicolumn{1}{c|}{time (s)}  & \multicolumn{1}{c|}{reward} & \multicolumn{1}{c|}{time (s)}  & \multicolumn{1}{c|}{reward} \\ \hline
\multirow{2}{*}{Bayes-POMCP} & 4x4                                       & \multicolumn{1}{l|}{0.47} & $46.75 \pm 6.03$                & \multicolumn{1}{l|}{1.96} & \bf{$49.08 \pm 6.32$}      \\ \cline{2-6} 
                             & 8x8                                       & \multicolumn{1}{l|}{0.79} & $79.75 \pm 13.51$                & \multicolumn{1}{l|}{3.24} & \textbf{$82.67 \pm 8.16$}      \\ \hline
\multirow{2}{*}{POMCP}       & 4x4                                       & \multicolumn{1}{l|}{0.13} & $42.5 \pm 6.8$                   & \multicolumn{1}{l|}{0.66} & $40.83 \pm 7.99$                \\ \cline{2-6} 
                             & 8x8                                       & \multicolumn{1}{l|}{0.15} & $18.33 \pm 31.61$             & \multicolumn{1}{l|}{0.71} & $35.33 \pm 33.32$              \\ \hline
\end{tabular}
\caption{Computational Analysis for POMCP variants with different grid sizes on the Frozen Lake Domain.}
\label{tab:comp_analysis}
\end{table}

\section{Study Domain}
We designed fourteen $8\times8$ grid maps for studying mixed-initiative human-robot teaming using the Frozen Lake domain. In the data collection study, we used one map for demonstrating the interface to the user, three maps as practice rounds for the user, and the remaining ten maps as part of the formal study. In the evaluation study, we used the same demo and practice maps and used six of the remaining ten maps (2 rounds per condition) for the user study. We chose to use $8\times8$ grid maps so that solving the task is neither trivial nor too complex, enabling the majority of the users to solve it within a reasonable timeframe. 

We designed each map such that there only exists one path to the goal. However, neither the robot nor the human is aware of this path due to the presence of slippery regions, which are only partially observable to each agent. Further, each map has the same number of human and robot errors in identifying slippery regions, but the number of steps to reach the goal can vary across maps.

\subsection{Heuristic Policy}
For the data collection study, we designed a simple heuristic policy for robot interventions to analyze how the style of intervention (e.g., interrupt or take-control) can influence team performance. The heuristic is a short-horizon planner that only evaluates whether executing the current human action will result in a dangerous state (i.e., a slippery region or hole) or will lead to a longer path ($> k$ steps) based on the robot's domain knowledge (Lines 4-9, Alg. 1). 

In Algorithm 1, the $\text{goal\_dist}$ refers to the distance to Manhattan distance to the goal from a given state. In Line 7, we compare whether the distance to the goal after executing the user's action is greater than the goal from a neighboring (\textbf{nbr}) state by $k$. If so, the robot intervenes. The heuristic employs a static intervention style (e.g., always interrupts). Further, the heuristic will renounce control to the user if they are persistent in executing the same action. 
\begin{algorithm}
\DontPrintSemicolon
\KwIn{$\text{World state}: x_t,  \text{Current human action}: a^H_t$, \text{prev\_interrupt} \codeComment{// True if robot intervened at $t-1$}}
$a^R_t \leftarrow \textbf{Not Intervene}$ \;
\If{\text{prev\_interrupt} == False}{
        $x_{t+1} \sim p(\cdot|x_t, a^H_t, a^R_t)$ \;
        \uIf{$x_{t+1} == \text{slippery}$ or $x_{t+1} == \text{hole}$}{
            $a^R_t \leftarrow \textbf{Intervene}$ \;
            $\text{prev\_interrupt} = \text{True}$ \;
        }
        \uElseIf{$\text{goal\_dist}(x_{t+1}) - \text{goal\_dist}(\textbf{nbr}(x_{t+1})) > k$}{
            $a^R_t \leftarrow \textbf{Intervene}$ \;
            $\text{prev\_interrupt} = \text{True}$ \;
        }
        \Else{
            $a^R_t \leftarrow \textbf{Not Intervene}$ \;
            $\text{prev\_interrupt} = \text{False}$ \;
        }
        
    }
    \Else{
        $\text{prev\_interrupt} = \text{False}$ \;
    }
    \textbf{return} $a^R_t, \text{prev\_interrupt}$\;
    \caption{Heuristic Policy for Robot Intervention}
\end{algorithm}

\section{Algorithm}
In this section, we present additional details of our algorithm implementation for mixed-initiative human-robot teaming.
\subsection{Implementation Details}
In the original POMCP algorithm as proposed by Silver and Veness \cite{silver2010monte}, the robot always receives an observation after performing an action in the environment. We adapt this to our mixed-initiative human-robot team setting, where the robot takes an action in response to the human action (i.e., the robot's observation). Hence, at the start of the interaction episode, we assume the human takes the first action, which forms the root node of the search tree (Alg. 2, Line 1). We perform several simulations (as determined by the search hyperparameter $n\_sims$) to select the best robot action.

At the start of the interaction, the root node is initialized with belief $b_0$ over user latent states. Since we assume no knowledge of the human, we set $b_0$ as beta particles representing a uniform distribution ($\beta(1,1)$). We follow the POMCP procedure to update the node statistics in each simulation. A key distinction between our approach and the regular POMCP is that we use the robot's estimation of the human latent state to simulate their behavior, as shown in the step function. In the case of the Frozen Lake domain, we only anticipate whether the human will comply or oppose the robot's intervention (detect or persistently move in the same direction after an intervention). If the user decides to oppose, we assume that it is equally likely for them to detect or move in the opposite direction. To anticipate multiple user action categories, we can use Dirichlet counts instead of beta priors to represent the belief.

To determine the best search parameters, we tested with simulated human models across different maps. For our user study, we set the following search parameters: $\gamma = 0.99,  \epsilon=\gamma^{30}, \text{n\_sims}=100$.
\subsection{Computational Analysis}
In this section, we report the average computation time taken to calculate the robot action at each timestep for solving a $4 \times 4$, and an $8 \times 8$ grid using variants of our proposed approach with simulated human models as shown in Table~\ref{tab:comp_analysis}. All experiments were conducted on an Alienware Aurora R13 Desktop PC with 12th Gen Intel Core i9-129000F (16 cores, 2.4 GHz) Processor. 

We report the average computation time for calculating the robot action at each step, and the total reward values averaged across three seeds and four simulated human models per map. The reward values were calculated using Equation 3 (in the Main paper) with max\_steps set to $50$ for the $4\times4$ map and $100$ for the $8\times8$ map. We note that for both the $4\times4$ and the $8\times8$ map, the total reward obtained by Bayes-POMCP for $100$ simulations is greater than POMCP with both $100$ and $500$ simulations. Further, the computation time for $100$ simulations with Bayes-POMCP is less than the computation time required with $500$ simulations with POMCP. Hence, in our user studies, we set $n\_sims = 100$, as it achieves good performance and requires less than a second to compute each robot action.

We note that for higher grid dimensions, the computational time required to select robot actions can be greater. In that case, we will need to make a tradeoff between computation time and performance by choosing a preset termination criteria for the MCTS procedure.

\begin{algorithm*}
\DontPrintSemicolon
\KwIn{$\text{Search Hyperparameters}: \gamma,  \epsilon, \text{n\_sims}$}
\SetKwBlock{SearchBlock}{function \textsc{Search}$(h)$:}{}
\SetKwBlock{RolloutBlock}{function \textsc{Rollout}$(s, h, a^R, {depth})$:}{}
\SetKwBlock{StepBlock}{function \textsc{Step}$(s, a^R)$:}{}
\SetKwBlock{SimulateBlock}{function \textsc{Simulate}$(s, h, depth)$:}{}
\begin{multicols}{2}
\SetAlgoNoLine\SearchBlock{
    \SetAlgoVlined
    \codeComment{ // ${b(h)}$ is the belief over augmented states (particle filter)}\;
    \For{$i \leftarrow 1 \cdots \text{n\_sims}$}{
        \codeComment{// Sample an augmented state $s$ from belief $b $}\;
       ${s} \sim {b(h)}$ \codeComment{// reference to a particle}\;
       \codeComment{// Copy to avoid changing the particle in root node}\;
       $\tilde{s} \leftarrow \textsc{Copy}({s})$\;
       \textsc{Simulate}$(\tilde{s}, h, 0)$\;
       $a^R \leftarrow \textsc{GreedyActionSelection}(h)$
    } 
    \textbf{return} $a^R$\;

    }
\textbf{end } \\
    \;

\SetAlgoNoLine \RolloutBlock{
    \SetAlgoVlined
    \If{$\gamma^{depth} < \epsilon$}{
        \textbf{return} 0 \;
    }

    $\color{skyblue} a^H, s' \leftarrow \textsc{Step}(s, a^R)$\;
    $h' \leftarrow (h, a^R, a^H)$\;
    $r \leftarrow \mathcal{R}(x,a^R,a^H)$

    $\tilde{a}^R \sim \textsc{Uniform}(h', \cdot)$\;
    \textbf{return} $r + \gamma \cdot \textsc{Rollout}(s', h', \tilde{a}^R, depth+1)$\;
}

\textbf{end} \\
\;

\SetAlgoNoLine \StepBlock{
    \SetAlgoVlined
    $(x, \chi) \equiv s$ \;
    $a^H \sim \chi$\;
    $x' \sim p(\cdot|x, a^R, a^H)$ \codeComment{// Assume world dynamics is known}\; 
    $\color{skyblue} \chi'_{s,a^R} \leftarrow \chi^{}_{s,a^R} + 1$ \codeComment{// Update Dirichlet (beta) counts}\; 
    \textbf{return} $a^H, (x', \chi')$\;
}
\textbf{end} \\
    \;

\SetAlgoNoLine \SimulateBlock{
    \SetAlgoVlined
    \If{$\gamma^{depth} < \epsilon$}{
        \textbf{return} 0 \;
    }
    \codeComment{// Update belief of current node}\;
    $b(h) \leftarrow b(h) \cup \{s\} \; \; \; $  \;
    
    \codeComment{// Robot Action selection based on current node statistics }\;
    $a^R \leftarrow \textsc{UCBActionSelection}(h)$\;
    
    \codeComment{// Check for termination after $a^R$ since robot can override $a^H$ }\;
    \If{$\textsc{IsTerminal}(ha^R)$}{
    $\textbf{return} \; \textsc{TerminalReward}(ha^R)$\;
    }

    \If{$ha^R \notin T$}{
        \codeComment{// Not Previously visited}\;
        $\textbf{return} \;  \textsc{Rollout}(s, h, a^R, depth)$\;
    }

    $\color{skyblue} a^H, s' \leftarrow \textsc{Step}(s, a^R)$\;
    $h' \leftarrow (h, a^R, a^H)$\;
    \If{$h' \notin T$}{             
        \codeComment{// Construct node $h'$ and add to $T$}\;
         $T(h') \leftarrow $  \textsc{ConstructNode($T, h'$)}\;
    }
    $r \leftarrow \mathcal{R}(x,a^R, a^H)$\;
    $R \leftarrow r + \gamma \cdot \textsc{Simulate}(s', h', depth+1)$\;
    
    \codeComment{// Update node statistics} \;
    $N(ha^R) \leftarrow N(ha^R) + 1$\;
    $V(ha^R) \leftarrow \frac{N(ha^R) - 1}{N(ha^R)}V(ha^R) + \frac{1}{N(ha^R)}R$\;
    $N(h') \leftarrow N(h') + 1$\;
    $V(h') \leftarrow \frac{N(h') - 1}{N(h')}V(h') + \frac{1}{N(h')}R$\;
    \textbf{return} $R$ \;
    
}
\textbf{end}
 \end{multicols}
 \caption{\bf Bayes-POMCP Search for Robot Action Selection in Mixed-Initiative Teams.}
  \label{algo:BayMax}
 \end{algorithm*}





\begin{table}[]
\fontsize{7.5pt}{9.5pt}\selectfont
\begin{tabular}{|c|c|c|c|c|c|c|}
\hline
Hypotheses & I.V.                                                                 & Levels & n  & D.V.                                                   & Effect Size & Power  \\ \hline
H1A        & \begin{tabular}[c]{@{}c@{}}Robot\\ Intervention\\ Style\end{tabular} & 5      & 29 & Reward                                                 & 0.201       & 0.3728 \\ \hline
H1B        & \begin{tabular}[c]{@{}c@{}}Robot\\ Intervention\\ Style\end{tabular} & 5      & 29 & \begin{tabular}[c]{@{}c@{}}User\\ Ranking\end{tabular} & 0.428       & 0.995  \\ \hline
H2A        & \begin{tabular}[c]{@{}c@{}}Robot\\ Policy\end{tabular}               & 3      & 25 & Reward                                                 & 0.738       & 0.923  \\ \hline
H2B        & \begin{tabular}[c]{@{}c@{}}Robot\\ Policy\end{tabular}               & 3      & 25 & \begin{tabular}[c]{@{}c@{}}User\\ Ranking\end{tabular} & 0.528       & 0.581  \\ \hline
\end{tabular}
\caption{Power and Effect size analysis}
\label{tab:power}
\end{table}

\section{Simulated Human Experiments}
We first validate our proposed approach with simulated human models before deploying it on real users in the human-subjects experiment. For our analysis, we consider two latent parameters to modulate simulated human behavior, namely expertise, $\psi$, and compliance, $\theta$. We model simulated humans such that the latent parameters are either static or dynamic during the interaction. 

\begin{figure}[t]
    \centering
    \includegraphics[width=0.8\columnwidth]{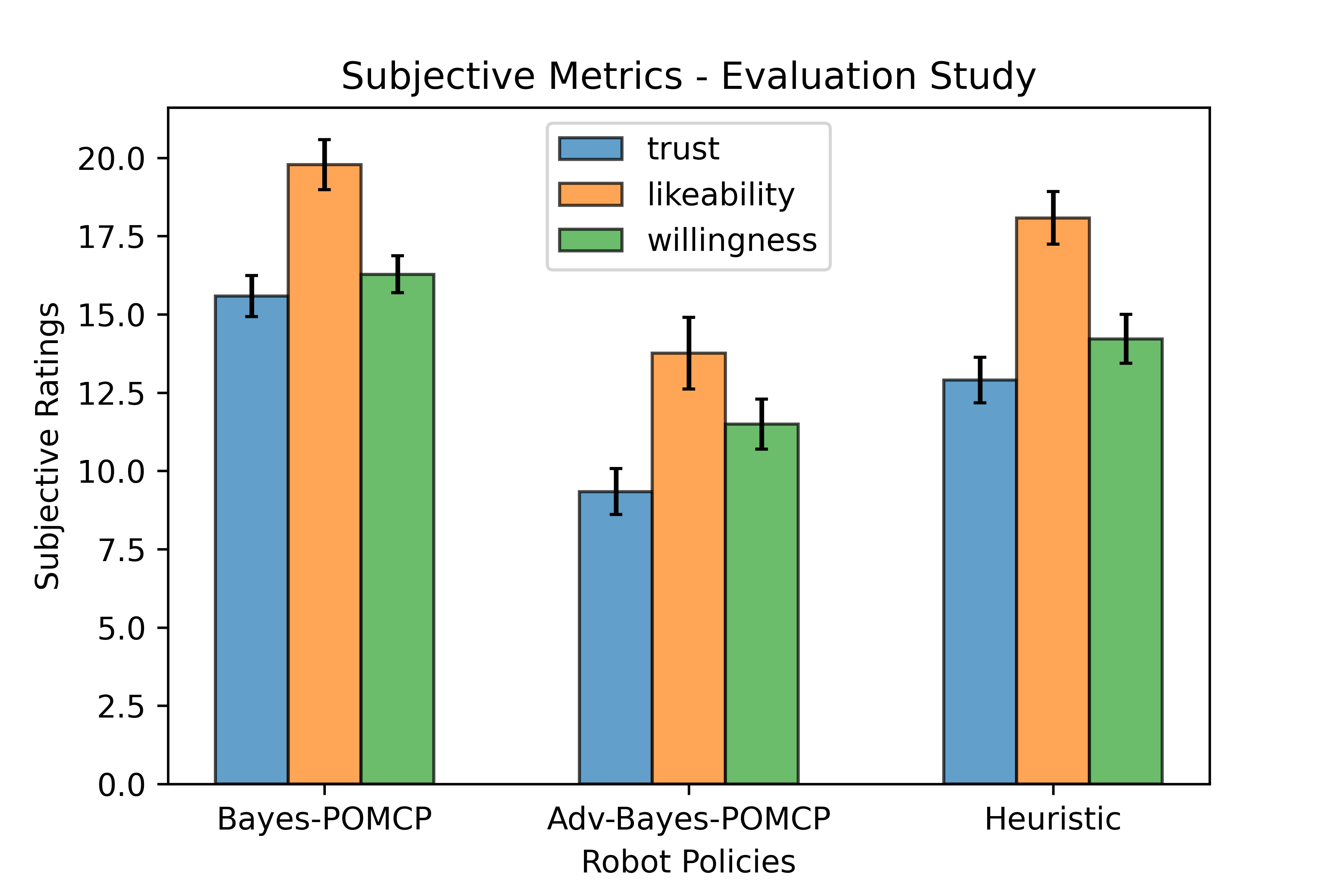}
    \caption{Self-reported measures: Evaluation Study.}
    \vspace{-1.5em}
    \label{fig:eval}
\end{figure} 

\subsubsection{Static Users}
\label{sec:static_users} We first test our models with a population of static users, whose latent parameters -- expertise ($\psi$) and compliance ($\theta$) are fixed. In the Frozen Lake domain, all users are suboptimal as they only have partial knowledge of the slippery regions, which is kept consistent across all simulated human models. The users' expertise doesn't correlate with their knowledge of slippery regions but instead reflects their ability to find a path from their current location to the goal based on their limited knowledge of these slippery areas. We utilize $\epsilon-$greedy policies to determine the users' planning, where the greedy policy is an A* search to find the optimal path to the goal. The users' expertise $\psi$ can range between $0-1$ and is the inverse of the $\epsilon$ value in the $\epsilon$-greedy policy, i.e., $\psi = 1 - \epsilon$.

The users' compliance $(\theta)$ refers to their tendency to comply with the various robot intervention styles. Users with higher compliance tend to detect less and rely more on the robot. Conversely, users with lower compliance rates tend to exercise caution, use the detection sensor more frequently, and might actively resist the robot by moving in the opposite direction when the robot intervenes. We model a population of users with varying compliance rates drawn from different $\beta$ distributions $\{ \beta(20,80)$; $\beta(40,60)$; $\beta(50,50)$; $\beta(60,40)$; $\beta(80,20)\}$. 

\subsubsection{Dynamic Users}
\label{sec:dynamic_users} In the next set of experiments, we only modify the compliance rates during the interaction and set the capability $\psi$ as fixed ($=0.7$) since we found that this was the closest to the scores obtained by real users in the data collection study. Users' rate of compliance is inherently dependent on their trust in the robot, which is a dynamic parameter \cite{chen2018planning}. Hence we chose to model a population of users whose compliance rates change based on their interaction history. From the first user study, we know that most users do not prefer the take-control agent. Hence, we model some individuals whose compliance decreases with robot policies that excessively take control and increase for the conditions where the robot provides explanations (i.e., we assume that $\theta$ is correlated with their preferences). We also model a subset of users whose compliance rates increase gradually with successful collaboration and reduce after failing (e.g., slipping into holes).
\section{Statistical Analysis}
\subsection{Study Questionnaires}
We list the pre- and post-experiment questionnaires used in both the human-subjects experiments in Table~\ref{tab:surveys}. We also show the user's self-reported trust, likeability and willingness to comply measures for the different robot policy conditions in the evaluation study in Figure~\ref{fig:eval}. Error bars indicate standard error. Trust and willingness to comply were measured via a 4-item Likert scale, and robot likeability was measured on a 5-item semantic continuum.
\begin{table*}
     \begin{subtable}[h]{0.45\textwidth}
        \fontsize{7.5pt}{9.5pt}\selectfont
\centering
\begin{tabular}{lll}
\hline
 & Post Trial Trust Questionnaire & \multicolumn{1}{l}{} \\ \hline
1. & I think the robot's behavior can be predicted from & \\ & moment to moment. &  \\
2. & I can count on the robot to do its job. &  \\
3. & I have faith that the robot will be able to cope with & \\ & similar situations in the future. &  \\
4. & Overall, I trust the robot. &  \\ \hline
\end{tabular}
\vspace{1mm}
\caption{ Trust Questionnaire (From Muir's Trust Questionnaire \cite{muir1989operators}) -- administered after each round in both user studies}
     \end{subtable}
    \hfill
\begin{subtable}[h]{0.45\textwidth}
        \fontsize{7.5pt}{9.5pt}\selectfont
\centering
\begin{tabular}{lll}
\hline
 & \textcolor{white}{AAAAAAAAA} Robot Likeability  \textcolor{white}{AAAAA} & \multicolumn{1}{l}{} \\ \hline
1. & Dislike & Like \\
2. & Unfriendly & Friendly \\
3. & Unkind & Kind \\
4. & Unpleasant & Pleasant \\
5. & Awful & Nice \\ \hline
\end{tabular}
\vspace{1mm}
\caption{Robot Likeability (From Godspeed Questionnaire \cite{bartneck2009measurement}) -- administered after each round in both user studies}
     \end{subtable}
    \hfill

    \begin{subtable}[h]{0.45\textwidth}
    \fontsize{7.5pt}{9.5pt}\selectfont
\centering
\begin{tabular}{lll}
\hline
 & Willingness to Comply  & \multicolumn{1}{l}{} \\ \hline
1. & I would be willing to improve my decision after the &  \\
& robot's intervention. & \\
2. & I would be willing to invest a lot of effort in making &  \\ & another decision. & \\
3. & The robot's intervention makes me willing to do a &  \\
&  better job of making better decisions.  & \\
4. & The robot's intervention provides suggestions as & \\ & to how I could make better decisions. &  \\

\hline
\end{tabular}
\vspace{1mm}
\caption{User's willingness to comply or change their decision after robot's interventions (adapted from a survey on human-human interactions \cite{raemdonck2013feedback})}
     \end{subtable}
          \hfill
    \begin{subtable}[h]{0.45\textwidth}
        \fontsize{7.5pt}{9.5pt}\selectfont
\centering
        \begin{tabular}{ll}
\hline
 & Negative Attitude Towards Robots  \\ \hline
1. & I would feel relaxed when talking with robots. \\
2. & I would feel uneasy if robots really had emotions. \\
3. & Something bad might happen if robots developed into living beings. \\
4. & I would feel uneasy if I was given a job where I had to use robots. \\
5. & If robots had emotions, I would be able to make friends with them. \\
6. & I feel comfortable being with robots that have emotions. \\
7. & The word "robot" means nothing to me. \\
8. & I would feel nervous operating a robot in front of other people. \\
9. & I would hate the idea that robots or artificial intelligence. \\
& were making judgements about things. \\
10. & I would feel very nervous just standing in front of a robot. \\
11. & I would feel that if I depend on robots too much, \\ & something bad might happen. \\
12. & I would feel paranoid talking with a robot. \\ 
13. & I am concerned that robots would be a bad influence on children. \\
14. & I feel that in the future, society will be dominated by robots. \\ \hline
\end{tabular}
\vspace{1mm}
\caption{Negative Attitude Towards Robots Scale (NARS) -- Pre-Study Questionnaire}
    \end{subtable}
     \caption{Subjective Metrics used in both human-subjects experiments}
     \label{tab:surveys}
     
\end{table*}

\subsection{Power and Effect Size Analysis}
For all our hypotheses, \textbf{H1A, H1B, H2A, H2B}, we used non-parametric tests as the dependent variable was ordinal data (for \textbf{H2A}, \textbf{H2B}) or the model did not pass the assumptions needed for parametric tests (for \textbf{H1A}, \textbf{H1B}). We report the effect size and statistical power for our analyses in Table~\ref{tab:power}. DV and IV in the table refer to the dependent and independent variables in each hypothesis, respectively, and $n$ refers to the number of subjects (both experiments followed a within-subjects design).

\subsection{Statistical Model Assumptions}
We conduct three repeated measures ANOVA to assess whether the self-reported metrics (trust, likeability, and willingness to comply) are significantly dependent on the robot policy in the Evaluation study. The subjective metrics were the D.V., and robot policy, number of rounds completed, demographics (age, gender, prior robotics experience), and
pre-study questionnaire responses of the user as the I.V. to build a multi-linear regression model. We ensured that each of the models passed the required assumptions for ANOVA (normality using Shapiro-Wilk's test and homoscedasticity using the Levene's test). We eliminate effects backward and report the significance of the final model with the lowest AIC score using the ANOVA and Tukey HSD. For trust and willingness to comply, our Bayes-POMCP approach is significantly higher than heuristic $(z=4.19, p<.001)$, and the heuristic is significantly higher than Adv-Bayes-POMCP $(z=7.93, p<.001)$. For robot likeability Bayes-POMCP and the heuristic have similar scores (not significant) and are better than the Adv-Bayes-POMCP.

\bibliographystyle{plain}
\bibliography{references.bib}
